\documentclass[10pt,twocolumn,letterpaper]{article}

\usepackage{wacv}
\usepackage{times}
\usepackage{epsfig}
\usepackage{graphicx}
\usepackage{amsmath}
\usepackage{amssymb}
\usepackage{booktabs}
\usepackage{subcaption}
\usepackage{amssymb}
\usepackage{pifont}
\newcommand{\cmark}{\ding{51}}%
\newcommand{\xmark}{\ding{55}}%

\usepackage[nice]{nicefrac}
\usepackage[pagebackref=true,breaklinks=true,letterpaper=true,colorlinks,bookmarks=false]{hyperref}

\wacvfinalcopy
\def\wacvPaperID{738}

\begin{document}

\title{Disentangling Human Dynamics for \\Pedestrian Locomotion Forecasting with Noisy Supervision}

\author{Karttikeya Mangalam$^{1,3}$, Ehsan Adeli$^1$, Kuan-Hui Lee$^2$, Adrien Gaidon$^2$, Juan Carlos Niebles$^1$\\
$^1$Stanford University \quad $^2$Toyota Research Institute \quad $^3$University of California, Berkeley \\
\tt\small mangalam@cs.berkeley.edu \{eadeli, jniebles\}@cs.stanford.edu \\ \tt\small \{kuan.lee, adrien.gaidon\}@tri.global
}
\maketitle
\begin{abstract}
We tackle the problem of Human Locomotion Forecasting, a task for jointly predicting the spatial positions of several keypoints on human body in the near future under an egocentric setting. In contrast to the previous work that aims to solve either the task of pose prediction or trajectory forecasting in isolation, we propose a framework to unify these two problems and address the practically useful task of pedestrian locomotion prediction in the wild. Among the major challenges in solving this task is the scarcity of annotated egocentric video datasets with dense annotations for pose, depth, or egomotion. To surmount this difficulty, we use state-of-the-art models to generate (noisy) annotations and propose robust forecasting models that can learn from this noisy supervision. We present a method to disentangle the overall pedestrian motion into easier to learn subparts by utilizing a pose completion and a decomposition module. The completion module fills in the missing key-point annotations and the decomposition module breaks the cleaned locomotion down to global (trajectory) and local (pose keypoint movements). Further, with Quasi RNN as our backbone, we propose a novel hierarchical trajectory forecasting network that utilizes low-level vision domain specific signals like egomotion and depth to predict the global trajectory. Our method leads to state-of-the-art results for the prediction of human locomotion in the egocentric view. Project page: \href{https://karttikeya.github.io/publication/plf/}{https://karttikeya.github.io/publication/plf/}
\end{abstract}

\section{Introduction}
Pedestrians are one of the most vulnerable and prevalent entities in self-driving scenarios. The ability to predict their dynamics in the near future can assist in making proper decisions for immediate next action the vehicle needs to take. Forecasting human locomotion is useful in several downstream tasks for self-driving cars such as reasoning about pedestrian intent, path planning and reactive control. It is also relevant for social robots that require anticipating future human movements for collision avoidance and navigation. Human dynamics or locomotion can be defined in terms of the joint spatial movement of several keypoints on the human body. Forecasting future human dynamics can lead to foretelling certain undesirable activities like falling, which can then be planned for. It is the final product of a complex interaction between large scale trajectorial motion and finer body limb movements. In previous works, often one of these two critical components is studied. In this work, we propose to predict human locomotion by disentangling the global and local components. An additional challenge to this task is the scarcity of human annotated pedestrian pose datasets in egocentric view. To this end, we use off-the-shelf estimation models to generate noisy ground-truth data for training our model.

 \begin{figure}
\begin{center}
\includegraphics[width = \linewidth]{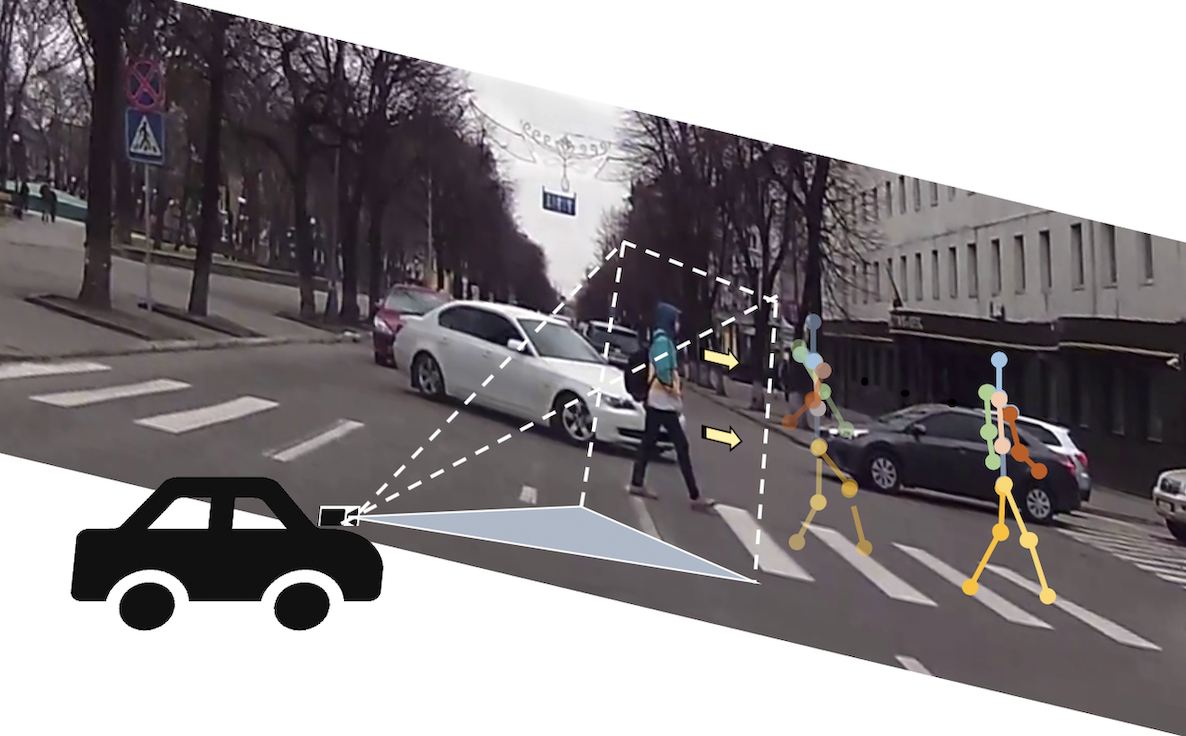}
\end{center} \vspace{-20pt}
\caption{Egocentric pedestrian locomotion forecasting. Locomotion is defined as the overall motion of keypoints on the pedestrian in contrast to predicting just the position (trajectory prediction) or the pose (pose forecasting).}
\label{fig:pull}
\end{figure} 

Developing computational methods for modeling human dynamics and forecasting how the pose might change in the future is itself an extremely challenging task. The first level of complexity comes from the inherent multimodal nature of pedestrian locomotion distribution. The space of possible future locomotion is both complex and uncertain even conditioned on the observed history. Furthermore, in real scenarios, the pedestrians often occlude with each other or other objects in the scene. Moreover, obtaining full annotations of the dynamics (trajectory and pose) is a very intensive task. Therefore, in contrast to previous works \cite{chiu2018action,martinez2017human,jain2016structural,alahi2016social,huang2019uncertainty}, applying fully-supervised methods is hardly possible for real-world in-the-wild applications. 

\begin{figure*}
\begin{center}  \vspace{-6pt}
\includegraphics[width = \textwidth]{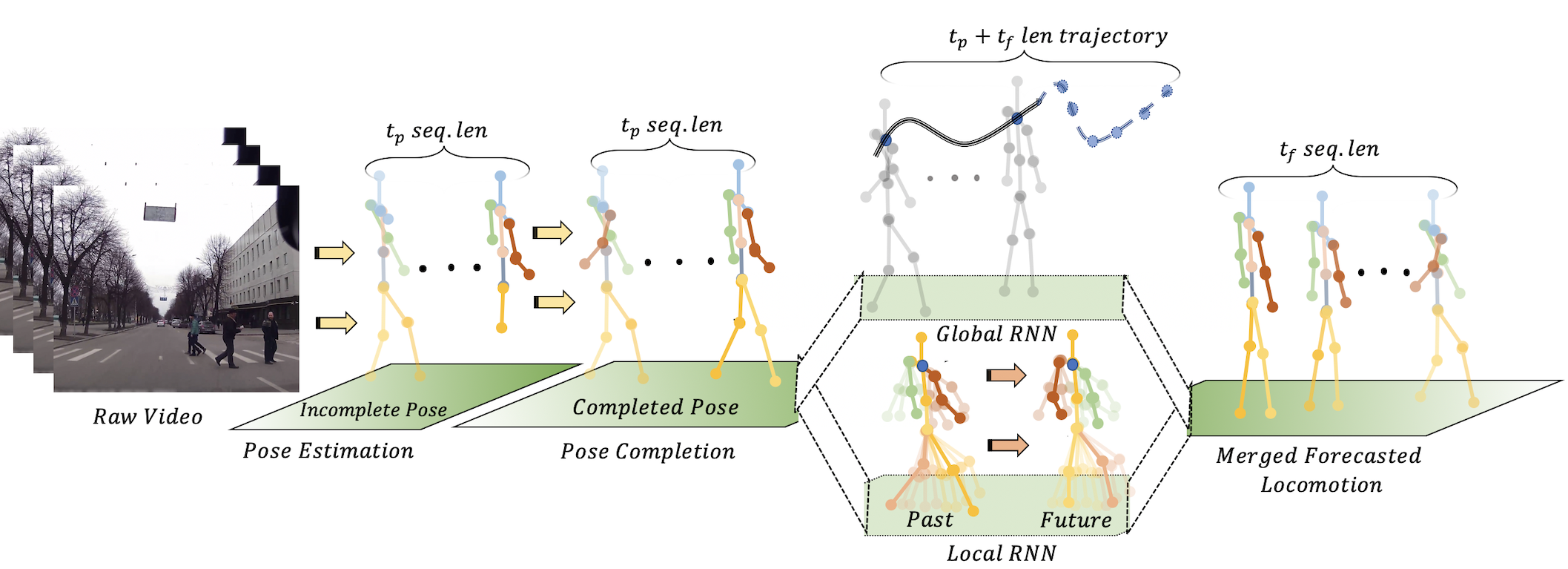}
\end{center} \vspace{-20pt}
   \caption{An illustration of the proposed method for human locomotion forecasting with noisy supervision. The ``Raw Pose" plane represents the noisy input pose sequence with missing joint detection (Section \ref{loose_supervision}). The ``Complete Pose" plane denotes the output from the pose completion module with filled in joint positions (Section \ref{posecompletion}). The completed pose is then split into the global and local streams (Section \ref{decomposition}) which separate concurrent motions. The prediction modules forecast the future streams as outlined in Section \ref{forecasting}. Finally, these streams are merged to predict future pedestrian locomotion. 
   \label{fig:system}
   }
   \vspace{-6mm}
\end{figure*}

In this work, we address the challenge of predicting human locomotion in the wild without explicit human labelled ground-truth supervision. We start with noisy machine generated supervision for forecasting using off the shelf pose detectors like OpenPose \cite{cao2017realtime, wei2016cpm, cao2018openpose}. We then propose a forecasting pipeline (Fig. \ref{fig:pull}) to complete the missing detections and somewhat denoise the detected poses. Furthermore, we identify the complexities associated with learning to forecast human locomotion in the wild in an end-to-end fashion and introduce a disentangling method to overcome this problem. While almost all of the previous work for prediction of human dynamics \cite{chiu2018action,martinez2017human,ma2017forecasting, katircioglu2018learning} target the problem of predicting global (\ie, trajectories \cite{Vemula-2018-107260, kooij2019context, kooij2014analysis}) and local movements (\ie, human pose and joints motion \cite{chiu2018action, jain2016structural}) separately, we propose to unify the previous works on future pedestrian pose and trajectory prediction under the single problem of locomotion forecasting. To achieve this goal, we propose a sequence-to-sequence pose forecasting model (see Fig. \ref{fig:system}) that build upon this decomposition. Lastly, we show that with these advancements our proposed method achieves better results than several other previous works and confirm our hypothesis of disentangling human dynamics for pedestrian locomotion forecasting in the wild.

In summary, our contributions are two-fold. \textbf{First}, we propose the pose completion and decomposition modules that uses \textit{disentanglement of global and local motion} to reduce the overall forecasting complexity and enables learning under noisy supervision. \textbf{Second}, we present the pose prediction module consisting of a novel egocentric trajectory prediction network based on the Encoder-recurrent-Decoder architecture that utilizes domain specific settings of egomotion to forecast these different granularity streams. They are merged for the final locomotion forecasting.

\section{Related Work}
\noindent\textbf{Trajectory Prediction:}
The literature for trajectory prediction can be split into two major categories of work: (1) works that aim to predict human trajectories from the points of view of either other humans in the scene (human-human) \cite{alahi2016social,yagi2018future,gupta2018social,sadeghian2018sophie} or from a top down view \cite{alahi2016social,gupta2018social,sadeghian2018sophie, zou2018understanding, 2017arXiv171004689V}; and (2) works that predict the paths of other vehicles on the road from the viewpoint of a car (car-car)  \cite{huang2019uncertainty,lee2017desire}. 

The previous work on human-human trajectory prediction often solves the problem from a top-down prospective \ie, \textit{bird's eye view}. For instance, Social LSTM \cite{alahi2016social} jointly predicted the paths of all the people in a scene by taking into account the common sense rules and social conventions that humans typically utilize as they navigate in shared environments. Social GAN \cite{gupta2018social} defined a recurrent sequence-to-sequence model using Generative Adversarial Networks \cite{goodfellow2014generative} (GAN) that observes motion histories to predict future behavior with spatial pooling thereby tackling the multi-modality associated with motion prediction. In \cite{zou2018understanding}, Zou et al. proposed to infer the latent factors of human decision-making process in an unsupervised manner by extending the Generative Adversarial Imitation Learning framework to anticipate future paths of pedestrians (human-car) from the top-down view. In \cite{sadeghian2018sophie}, SoPhie introduces an attentive GAN that predicts individuals' paths by taking into account physical constraints, \ie, scene context and other peoples' trajectories. Several work model the human trajectories with social forces \cite{helbing1995social} and attraction repulsion models using functional objects \cite{6751387}. Also, approaches using probabilistic modelling of trajectories with Gaussian processes \cite{ellis2009modelling} and mixture models \cite{wiest2012probabilistic} have been employed. Among the few works for trajectory prediction in the human's \textit{egocentric view}, \cite{yagi2018future} predicted trajectories of other pedestrians in an egocentric setting with a tri-stream CNN architecture.

Car trajectory prediction from \textit{egocentric view} were done by explicitly modeling other agents (cars) along with motion with respect to the camera. Yao \etal~ \cite{yao2018egocentric} proposed a multi-stream recurrent neural network (RNN) encoder-decoder model for predicting vehicle location and scale using pixel-level observations. 
DESIRE 
\cite{lee2017desire} was proposed to generate several possible trajectories for vehicles using both scene and vehicles, which were then ranked in an Inversion of Control (IOC) framework to choose the most likely one. Similarly, Huang \etal~\cite{huang2019uncertainty} introduced an uncertainty-aware method for prediction of self-trajectories. 

One of the most crucial tasks for self-driving car or social robots is to predict future dynamics of pedestrians (\ie, car-human task). However, recent work focuses on human-human or car-car prediction, both of which domains lack either egocentric cues like  monocular depth cue and large camera motion due to ego-vehicle or pedestrian centered features like human pose and intention.

\noindent\textbf{Pose Forecasting}: 
A number of recent work predicted 3D poses using fully annotated data with exact 3D positions. These works included Structural RNN \cite{jain2016structural}, convolutional sequence-to-sequence model with encoder-decoder architecture \cite{li2018convolutional}, and human motion prediction with adversarial geometry-aware based on geodesic loss \cite{gui2018adversarial}. 
Some other works introduced models that can predict poses in both 2D and 3D. Triangular-Prisms RNN (TP-RNN) \cite{chiu2018action} proposed an action-agnostic pose forecaster, and Matrinez \etal~\cite{martinez2017human} proposed a method based on residuals and a careful design of RNNs for human dynamics. These works concluded that zero-velocity is a very strong baseline for pose forecasting. Ghosh \etal~\cite{ghosh2017learning} introduced LSTM-3LR for long term forecasting and Zhai \etal~\cite{zhai2017learning} trained a model to forecast videos of human activities with multiple granularities which they use to forecast pose for future frame generation. 

\noindent\textbf{Denoising or Completing Poses:} Chang \etal~\cite{chang2017dr} propose a Denoising and Reconstruction Networks (DR-Net) for 3D human pose estimation from monocular RGB videos. Ballard \etal~\cite{ballard1987modular} leverage autoencoders for structured prediction of the human pose. Tekin \etal~\cite{tekin2016structured} use over complete autoencoders to account for dependencies between joints and use a latent space for 3D pose prediction from monocular images. Trumble \etal~\cite{trumble2018deep} introduce a convolutional symmetric autoencoder with reconstruction loss to encode skeletal joint positions, which simultaneously learns a deep representation of volumetric body shape.  

In contrast to the previous works, our solution for locomotion forecasting is different in three main aspects. \textit{First}, we focus on the egocentric setting for modelling pedestrians from the vehicle's view utilizing domain specific cues whereas the previous works are either on predicting other vehicle's movements \cite{huang2019uncertainty,lee2017desire} or for forecasting human trajectories from the top-down view \cite{alahi2016social,sadeghian2018sophie, 2017arXiv171004689V, gupta2018social}. \textit{Second}, all previous works focus on either predicting the pose \cite{chiu2018action,martinez2017human,jain2016structural} in the near future with small overall movement or on trajectory prediction \cite{alahi2016social,yagi2018future} whereas we propose a method to unify both for complete human locomotion forecasting. \textit{Third}, while almost all previous work is strongly supervised with human annotated ground-truth, our work is conducted in the wild using noisy supervision from off-the-shelf methods (Section \ref{loose_supervision}), which transfers the focus from expensive large quantity problem specific annotations to robust learning algorithms. 
 
\section{Proposed Method}
We frame the task of forecasting human locomotion in egocentric view (of the vehicle) as a sequence-to-sequence problem. Suppose $p_{t}$ denotes the human pose at time $t$, which contains $d$ two-dimensional joints, \ie, $p_{t} = \{(u^i,v^i)\}_{i=1}^d = \{\vec{u}_i\}_{i=1}^d$. Forecasting human locomotion is then defined as: given $t_{\text{p}}$ previous poses, $\{p_i\}_{i=t - t_p + 1}^{t} \equiv \{ p_{t - t_p + 1}, \hdots ,p_{t}\}$, predict the position of each of these joints for $t_f$ future frames. In other words, predict the sequence $ \{p_i\}_{i=t+1}^{t_f} \equiv \{ p_{t+1}, \hdots ,p_{t + t_f}\}$. Our method is illustrated in Fig.\ref{fig:system}. We start with noisy pose estimates generated using state of the art pretrained models (Section \ref{loose_supervision}). The missing and low confidence joints are filled in using the pose completion module (Section \ref{posecompletion}). The completed poses are then split into global and local streams (Section \ref{decomposition}) that are forecasted using Quasi Recurrent Neural Networks \cite{bradbury2016quasi} (Section \ref{QRNN}) in an Encoder-Recurrent-Decoder pose prediction module (Section \ref{forecasting}). These predictions combine to forecast the future pedestrian locomotion. 

\subsection{Machine Generated Noisy Supervision}
\label{loose_supervision}

\begin{figure}[]
\begin{center}
\includegraphics[width = 0.47\textwidth]{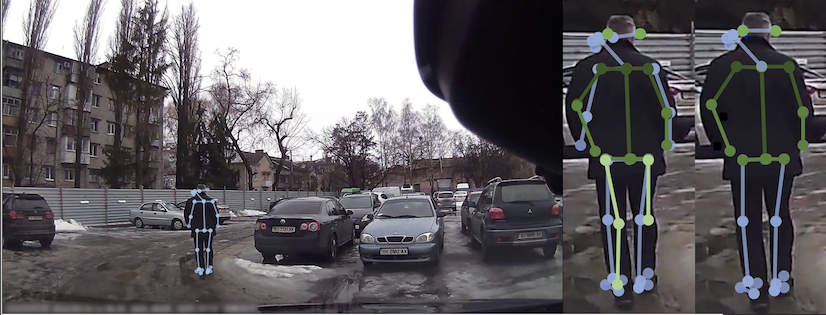}
\end{center} \vspace{-10pt}
   \caption{Qualitative results showing the results of pose completion (Section \ref{posecompletion}). From left to right: denoised pose, cropped version of the same pedestrian with the original OpenPose output (green), denoised pose (blue).}
   \label{fig: posecompletion}
\end{figure}

\noindent\textbf{Keypoint Detection} We use state-of-the-art models for \textit{multiple-person keypoint detection} module to autonomously generate dense but noisy frame-level supervision for human poses $p_t$. To the best of our knowledge, there are no existing datasets with manually annotated pose data that can be used for the task of human locomotion forecasting in the wild. So, we use pre-trained models on task-specific datasets to infer pose and depth on our datasets.  
\begin{figure*}
\vspace{-5mm}
\begin{center}
\includegraphics[width = \textwidth]{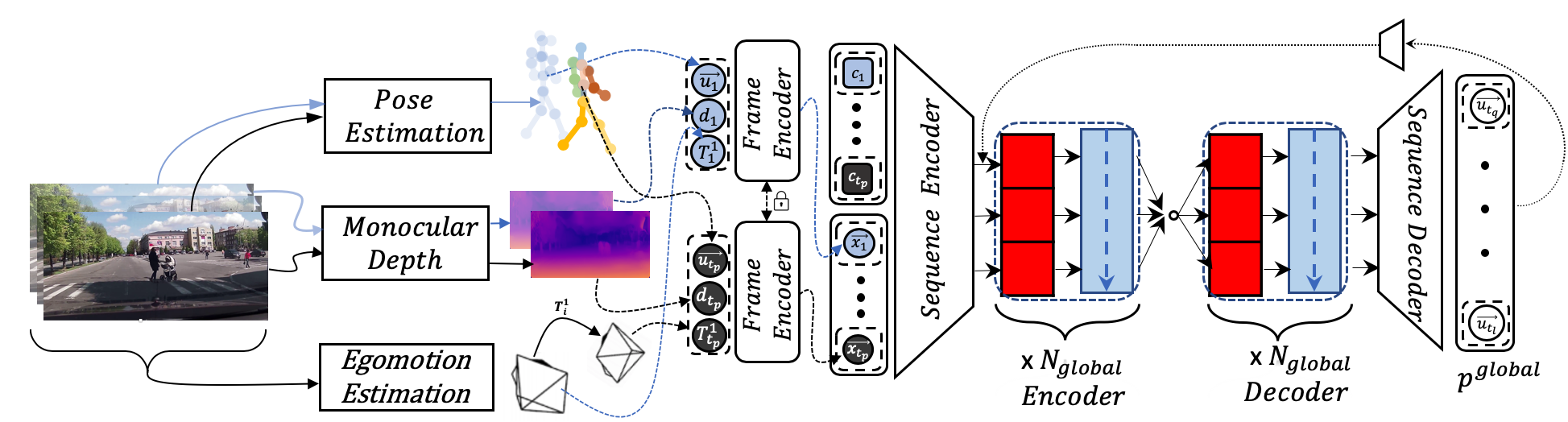}
\end{center}  \vspace{-22pt}
   \caption{Encoder-Recurrent-Decoder architecture for forecasting $p^\text{global}$. $\vec{u}_\alpha$, $d_\alpha$ and $c_\alpha$ represent coordinates, estimated monocular depth and confidence in joint $\tilde{i}$ at frame $\alpha$. $\mathcal{T}^\beta_\alpha$ represents the scene transformation matrix between frames $\beta$ and $\alpha$. The lock denotes the sharing of the frame encoder weights across different time steps of the input sequence. Dotted squares contain values concerned with the same frame.}
   \label{fig: global_prediction}
   \vspace{-2mm}
\end{figure*}
For each frame in the video, we create masked frames where everything except the pedestrians are masked out using the human labelled bounding box annotations. These masked frames are then processed through OpenPose \cite{cao2017realtime, wei2016cpm, cao2018openpose}, a pre-trained pose detection model to generate pose labels for every pedestrian in the frame. We notice that using masked full frames for generating frame-level pose supervision is faster and requires significantly lesser storage (upto $100$ times lesser) than processing each cropped out pedestrian one by one without any observable decrease in pose detection accuracy. For the $i^\text{th}$ joint detected in the $t^\text{th}$ frame, OpenPose detects the 2D coordinates $(u^i_{t},v^i_{t})$ and provides a confidence scores $c^i_t$, which is then used for denoising these detections as described in Section \ref{posecompletion}. These labelled keypoints form a human pose with $d = 25$ joints (Fig. \ref{fig: posecompletion}). 

\begin{figure}[t]
    \centering
    \begin{subfigure}[b]{\columnwidth}
        \centering
        \includegraphics[width=0.49\linewidth]{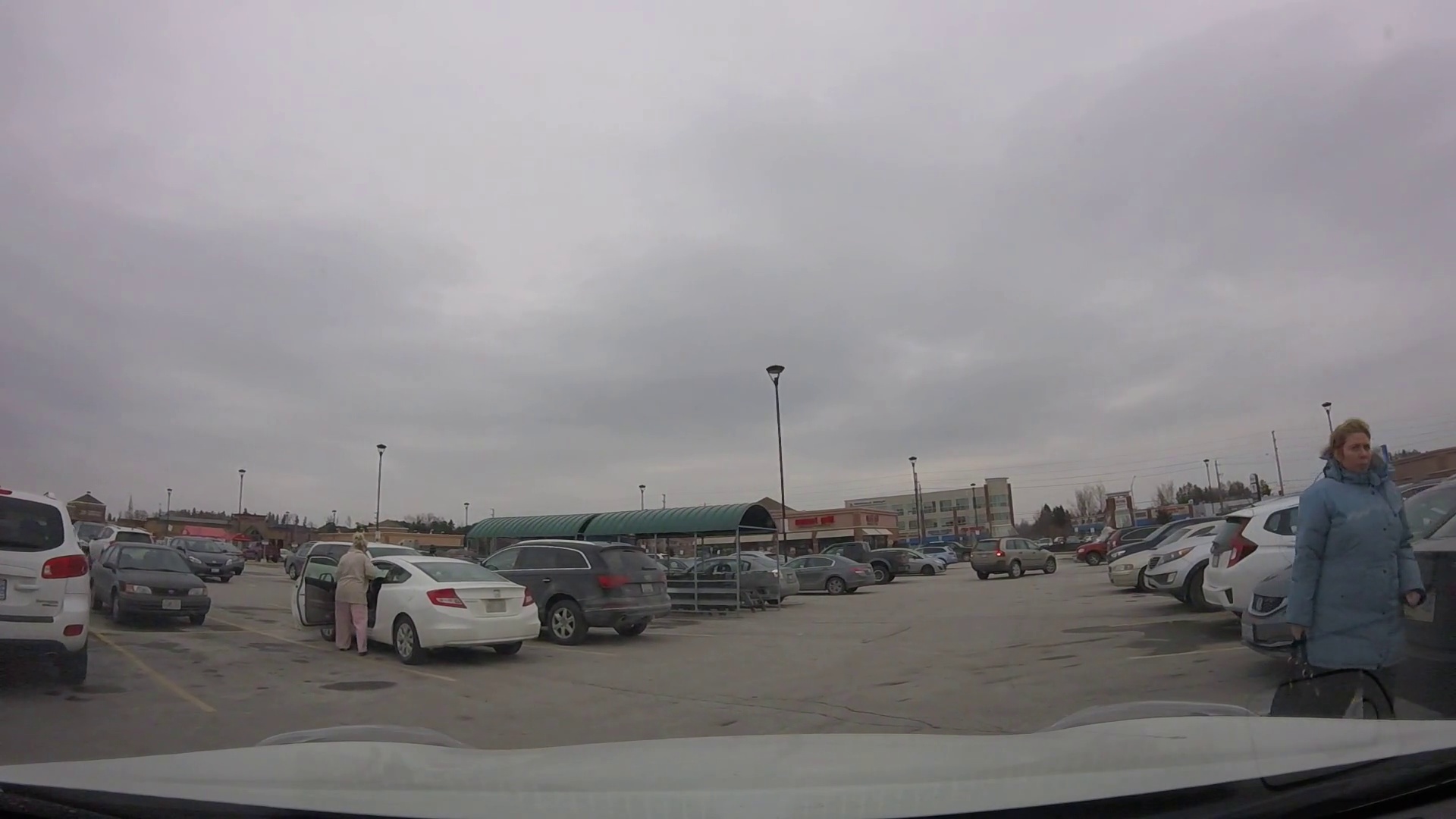}
        \includegraphics[width=0.49\linewidth]{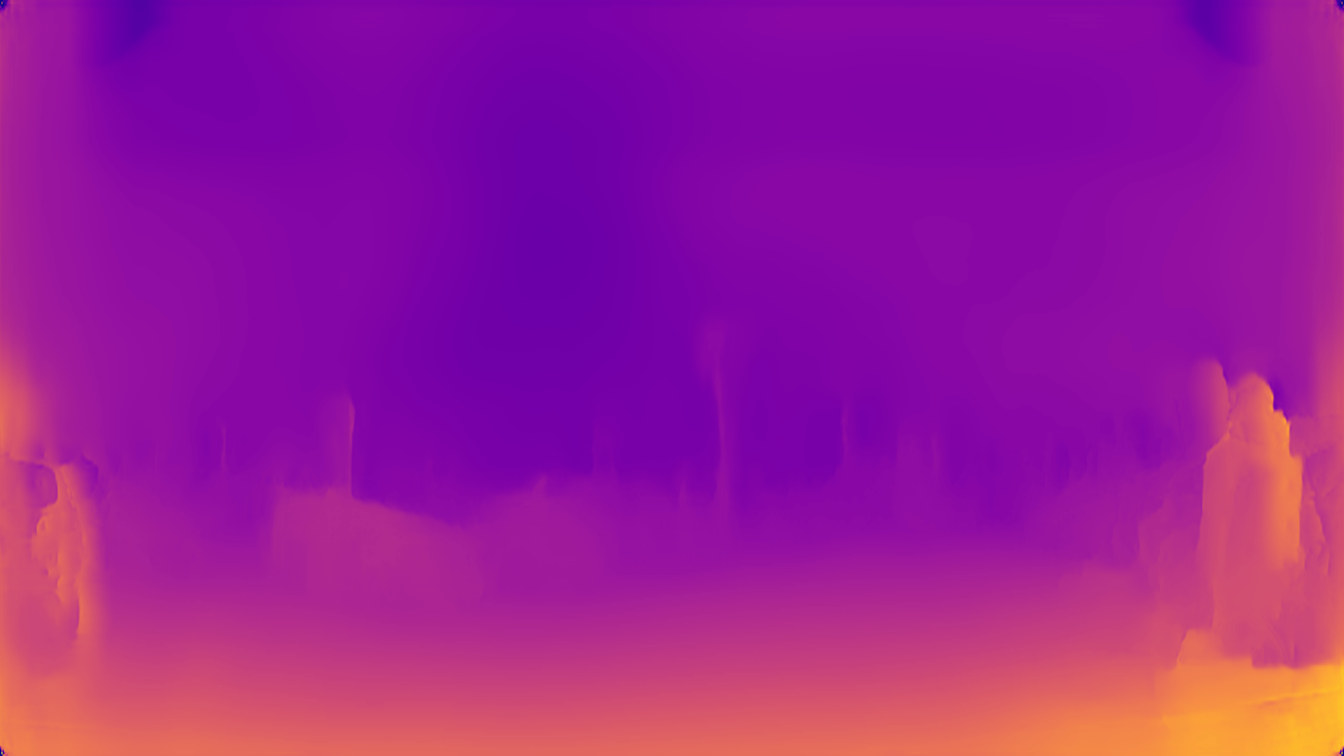}
        \vspace{-2mm}
    \end{subfigure}
    \caption{Examples of the input (left col.) and the output (right col.) of the monocular depth estimation module \cite{sudeep2018superdepth}.}
    \label{fig: superdepth}
    \vspace{-6mm}
\end{figure}

\noindent\textbf{Monocular Depth Estimation} We autonomously estimate depth in a monocular camera using SuperDepth \cite{sudeep2018superdepth}. SuperDepth extends a subpixel convolutional layer for depth super-resolution, where high-resolution disparities are synthesized from their corresponding low-resolution convolutional features. We train the depth estimation model in a self supervised fashion as proposed in \cite{sudeep2018superdepth}. Fig. \ref{fig: superdepth} shows qualitative examples of the depth estimation by SuperDepth. 

\noindent\textbf{Egomotion Estimation} We use the state-of-the-art-model unsupervised model \cite{zhou2017unsupervised} for autonomously estimating the camera motion that occurs between consecutive frames due to the movement of the egovehicle. However, since the method proposed in \cite{zhou2017unsupervised} can reliably estimate the scene transformation matrix $\mathcal{T}_i^j$ between frames $i$ and $j$ only frame very close to each other ($\Vert i - j\Vert \leq 5$  in a $30$ fps video) and the time horizons relevant for prediction are much larger ($\sim30$ frames), we chain the estimate between different overlapping short windows to obtain $\mathcal{T}_{i+k}^i$ for $k > 5$. As $k$ increases, this becomes another source of estimation noise that is taken into consideration while designing our proposed method. 

\subsection{Pose Completion}
\label{posecompletion}
Off-the-shelf pre-trained pose detectors often result in noisy detection of keypoints. In addition, the small scale and off co-planar orientation of pedestrians in real-world 2D videos, the pose detector may miss detecting some keypoints even for a completely visible occlusion-free pedestrians in the frame. Hence, we propose a pose completion network for completing the detected human poses obtained from OpenPose. This processing has a two-fold benefit. \textit{First}, it fills in the joints that are not detected by the pose detection module and so tackles the missing data problem. It also suppresses noise by filling in the low confidence output with better estimates. 
\textit{Second} and importantly, it enables us to decompose the motion (described in Section \ref{decomposition}) with noisy data. This is because otherwise separating the uncompleted global and local components of the motion would be perplexing as the joints flicker frequently.

\noindent\textbf{Network Architecture} We use an over-complete symmetric auto encoder similar to \cite{trumble2018deep, tekin2016structured}. The network is only trained on the subset of the total training data that has been assigned high confidence scores by Openpose. In other words, the poses are filtered by thresholding on OpenPose confidence scores to only include the examples with $c^i_t > \alpha_c, \forall\ i \in \{1, \hdots d\}$.
These high confidence examples are then used to train a symmetric autoencoder with dropout that embeds the $2d$-dimensional vectors $p_t$ to the latent dimension $d_{ae}$. Then, it maps them back to the denoised detections $\hat{p}_t$. Training with dropout in the input layer on high confidence examples, helps model the effect of missing data. Furthermore, supervising the loss function with the ground-truth on these good examples and the information bottleneck in the form of a narrow layer in the middle of network allows the model to learn to reconstruct the complete pose even if some of the joints are missing.  The training procedure details are in Section \ref{posecompletion_results}.

\noindent\textbf{Completion} The trained autoencoder is then used to estimate the full pose on all the low confidence noisy detections from the pose detection module. These estimates then fill in the missing and low confidence joint detections. Mathematically, $(u^i_t , v^i_t) \leftarrow (\hat{u}^i_t ,\hat{v}^i_t) \ \forall (i,t)$ such that $c^i_t \leq \alpha_c$.

\begin{figure}
\label{fig: com_decom}
\begin{center}
\includegraphics[width = 0.47\textwidth]{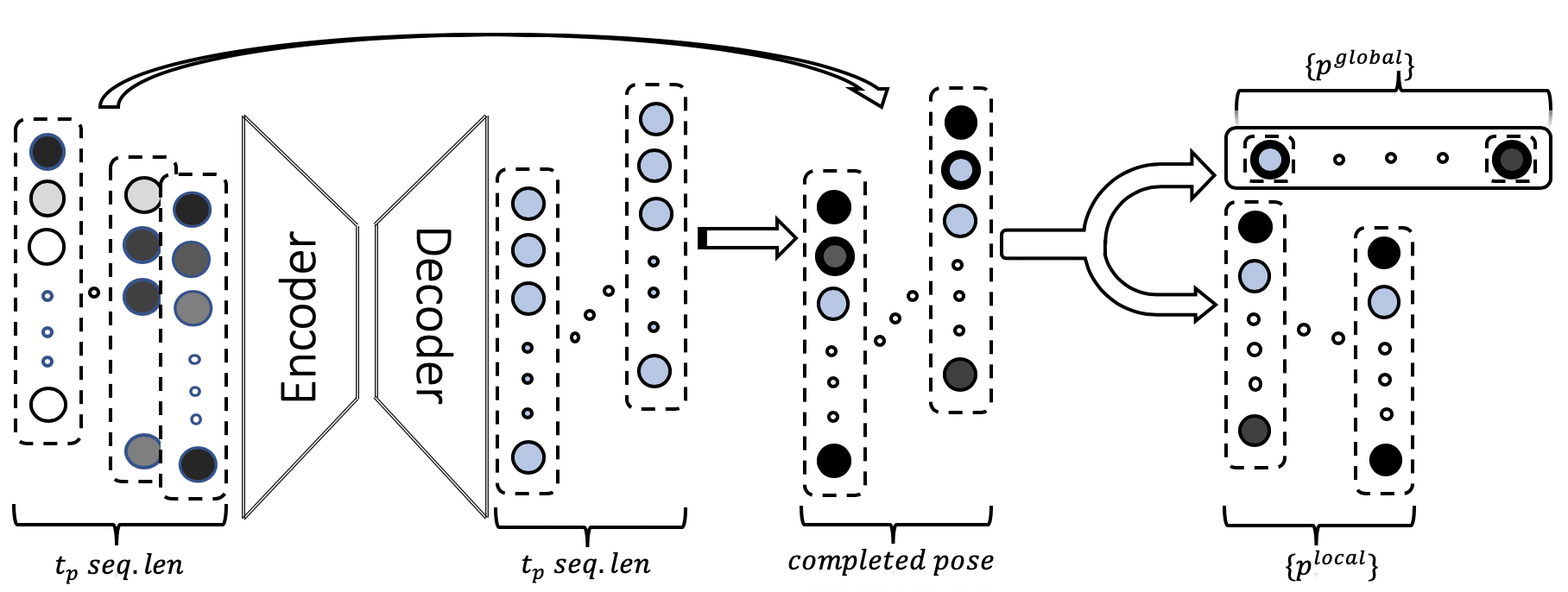}
\end{center} \vspace{-20pt}
   \caption{Architecture of the pose completion and disentangling module. The shades represent the confidence in locating the joint. Black represents highest confidence and white represents missing data. 
   All detections below confidence $\alpha_c$ are replaced with the autoencoder estimates (sky blue). It is then split into local and global streams for forecasting.
   }
   \label{fig:architecture}
   \vspace{-3mm}
\end{figure}
\subsection{Disentangling Pedestrian Locomotion into Global and Local Streams}
After noise suppression using the autoencoder and filling in the missing joints, we propose to disentangle the overall pose locomotion sequence $\{p_j\}_{j=t-t_p + 1}^{t}$ into two streams. A global motion stream encodes the overall rigid body motion of the pedestrian $\{p^{\text{global}}_j\}_{j=t-t_p + 1}^{t}$ and a local motion stream encodes the motion of the body with respect to the global stream $\{p^{\text{local}}_j\}_{j=t-t_p + 1}^{t}$. We hypothesize that these separate streams capture different granularities of simultaneous concurrent motion and need to be modelled and forecasted separately. The global motion stream $p^\text{global}$ models large scale movements of the pedestrian position with respect to the camera, such as those caused by the trajectory or egocentric motion of the camera mounted on the moving car \etc. The local motion stream captures the effect of depth change on the overall pose size and the movement of different joints of the pedestrian with respect to the global motion stream such small repetitive motions like swinging of arms. 

Our proposal to disentangle global and local motion is motivated by the relative difference in the nature and magnitude of motion exhibited by these streams. Consider for example, the actual path $\{p_j\}_{j=t-t_p + 1}^{t+t_f}$ traced by the elbow though space. In the original formulation this trajectory is quite complex, consisting of several swirls generated from gyrating around the shoulder and drifting with the rest of the body. In addition, this disentangling allows to significantly reduce the overall complexity, since each of the streams now model a much simpler and easier to predict motion. 

\begin{figure}
\begin{center}
\includegraphics[width = 0.47\textwidth]{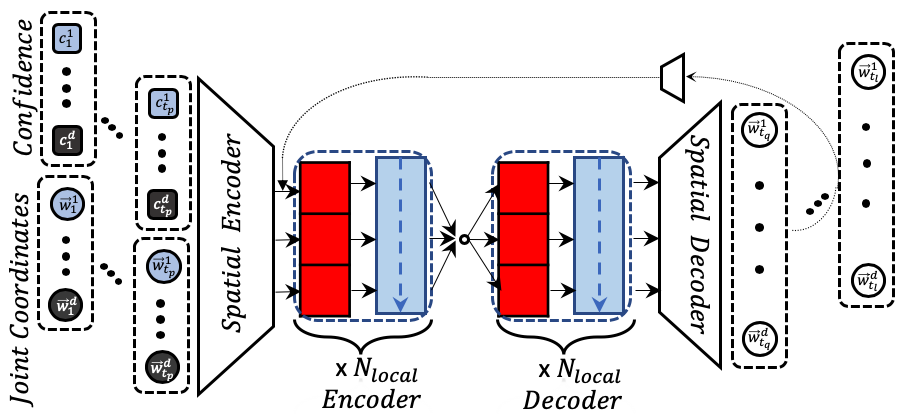}
\end{center} \vspace{-20pt}
   \caption{Architecture for forecasting the local stream, $p^\text{local}$. $\vec{w}^i_\alpha$ represent the relative coordinates of joint $i$ with respect to joint $\tilde{i}$ at frame $\alpha$ derived from the pose completion module.  Rest convention is same as Figure \ref{fig: global_prediction}.}
   \label{fig: local_prediction}
   \vspace{-1.5mm}
\end{figure}

We propose to use the neck joint sequence denoted by $({u}^{\tilde{i}}_j,v^{\tilde{i}}_j)_{j=t-t_p +1}^t$ as a representation of the global stream, because the neck is the most widely observed joint in the dataset. Note that this representation is possible only after filling in the missing data as discussed in Section \ref{posecompletion}. The local stream $p^{\text{local}}$ (denoted by $\vec{w}$) is then derived by reparameterizing the original stream $p_t$ (denoted by $\vec{u}$) as 
{
\begin{gather*}
\vec{w}^{\,i}_{t} = (u^i_{t},v^i_{t}) - (u^{\tilde{i}}_{t},v^{\tilde{i}}_{t}) 
\\
\forall i \in \{1, \hdots ,d\}, i \neq \tilde{i}, t \in \{t -t_p +1, \hdots,t\}  
\end{gather*}
\label{decomposition}}
\vspace{-15pt}
\subsection{Quasi-RNN Encoder-Decoder}
\label{QRNN}
The Quasi-Recurrent Neural Network \cite{bradbury2016quasi} forms the backbone of the sequence-to-sequence learning structure described in Section \ref{forecasting}. QRNNs consist of alternating convolutional and recurrent pooling module and is designed to parallelize efficiently better than vanilla LSTMs.  Similar to the findings of Bradbury \etal~\cite{bradbury2016quasi}, we found that Quasi-RNN trains faster ({$825$} ms/batch) compared to LSTM (${886}$ ms/batch) on a single GPU under same parameters and yield faster convergence for similar model capacity.

\noindent\textbf{Network Architecture} 
Our quasi-RNN encoder-decoder has $N$ layers of alternate convolutions and recurrent pooling, both in the input encoder and the output decoder. As illustrated in Figs. \ref{fig: global_prediction} and \ref{fig: local_prediction}, the recurrent pooling is a thin aggregation function applied to the convolutional activations. The encoder churns through the latent representation of the $t_p$ previous poses and encodes the necessary information into a context vector. This vector is then consumed by the QRNN decoder to forecast the future $t_f$ poses mapped back to the same latent space. 
The dotted arrow represents the use of the technique of teacher forcing \cite{williams1989learning} for training recurrent networks for human motion prediction, as also proposed by \cite{martinez2017human}.

\subsection{Forecasting the Future}
\label{forecasting}
After being split into separate granularities, the global and local streams are then forecasted separately with different prediction modules as described in the following. 

\noindent\textbf{Local Stream Forecasting}
As shown in Fig. \ref{fig: local_prediction}, the filled in and decomposed pose $\{p^\text{local}_j\}_{j=t-t_p+1}^{t}$ is used as the input to the pose prediction module. This module comprises of a spatial encoder with the latent dimension $d_{ae}$. The weights of this spatial encoder are separately trained using the autoencoder while the complexity of the latent space is similar.  The forecasting is processed in the latent space with $N_{\text{local}}$ layers of the QRNN Encoder Decoder module, briefly described in Section \ref{QRNN}. We use the latent space to forecast because as reported in \cite{tekin2016structured,katircioglu2018learning} and also as confirmed by our pose completion module experiments, the human pose lies on a low dimensional manifold because of the various kinematic constraints enforced by the human body. Thus, forecasting in this lower dimensional denser space makes the prediction easier for the quasi RNN module. The predicted latent pose is then mapped back into the image space with the spatial decoder to forecast $\{p^\text{local}_j\}_{t=t+1}^{j = t+t_f}$. 

\noindent\textbf{Global Stream Forecasting}
As mentioned in Section \ref{decomposition}, the global stream captures the effects of camera movement, perspective distortion from changing depth, and also the large-scale motion of target itself. Distortions due to the egocentric view, like perspective effects, can be implicitly learned using absolute coordinate trajectories while training. However, such an approach adds unnecessary complexity to the input distribution and hence is data inefficient. Instead, we propose to predict residuals from the first observed positions instead of forecasting absolute coordinates. In particular, we learn to predict the global stream from separately processed low level vision signals (monocular depth, camera egomotion). Furthermore, since these signals are machine generated and noisy, instead of using classical computer vision methods to employ these features like re-projecting the future positions with estimated camera pose matrices to cancel out egomotion effects, we propose a frame level encoder as depicted in Fig.\ref{fig: global_prediction}. The frame-level encoder is a feedforward neural network, which at sequence frame $\alpha$ takes as input observed coordinates ($\vec{u}^{\, \tilde{i}}_\alpha$), estimated monocular depth $d_{\alpha}$ of joint $\tilde{i}$, and a flattened $3 \times 4$ transformation matrix $T^1_\alpha$ estimating the change in the $6$D pose between start of the sequence and frame $\alpha$. The frame-level encoder architecture is motivated by the classical computer vision algorithm of projecting a point $(p_1,p_2,p_3)$ to a new point $(p'_1,p'_2,p'_3)$ under a transformation matrix $\mathcal{T}$ and hence is shared across different time steps. The output of encoder is a $2$ dimensional vector $\vec{x}_\alpha$, which is then pooled across time steps to be fed into the sequence level encoder. The rest of the architecture is similar to local stream forecasting module but with $N_\text{global}$ layers of QRNN. 

\noindent\textbf{Loss Metric} Both streams are trained with the $\ell_1$ loss between the predicted pose and the original pose (before completion) weighted by the confidence scores ${c^j_i}_{j=t+1}^{t+t_f}$ of OpenPose in the original detection. 
\subsection{Recombining the Streams}
These future predictions are finally recombined to forecast the overall human locomotion $\{p_j\}_{j=t}^{t+t_f}$ as follows:
{
\[
  \hat{p}_j =
  \begin{cases}
    p^\text{global}_j & \text{for } i= \tilde{i} \\
    p^{\text{global}}_j+  p^{\text{local,i}}_j &\text{for } i \neq \tilde{i}
  \end{cases} 
\]}$\forall j \in \{t+1, \hdots,t+t_f\}$ and $\forall i \in \{1, \hdots,d\}$. This operation forms the inverse of the splitting procedure (Section \ref{decomposition}).
\section{Experiments} 
In this Section, we run our method on a egocentric video dataset with pedestrian bounding-box annotations, and compare the results with widely used sequence-to-sequence and temporal convolution models. 

\noindent\textbf{JAAD Dataset} \cite{kotseruba2016joint}: Joint Attention in Autonomous Driving is a dataset for modelling attention in the context of autonomous driving scenarios. The dataset contains 346 videos with 82032 frames,  where the videos are the recorded with front-view camera under various scenes, weathers, and lighting conditions. It is also one of the very few autonomous driving datasets that has temporally dense pedestrian bounding box annotations. For example, popular egocentric datasets like KITTI \cite{geiger2012we}, Berkeley DeepDrive \cite{yu2018bdd100k}, and  CityScapes \cite{cordts2016cityscapes} do not contain human bounding boxes in a dense per-frame fashion. Similarly, PedX \cite{kim2019pedx} and Mapillary \cite{MVD2017} has pedestrian bounding box annotations, but these are separate images and not videos. 

The videos in JAAD range from short $5$ second clips to longer $15$ second videos shot through wide angle camera mounted inside the car in centre of windshield. It contains $64$ clips of resolution $1920 \times 1080$ pixels and rest $293$ of resolution $1280 \times 720$ pixels. Employing the procedure mentioned in \ref{loose_supervision}, a little under $1500$ non-overlapping $30$ frame length human locomotion examples are extracted which are then split into train and test sets while taking care not to have the same pedestrian's trajectories in both groups. 
\subsection{Experimental Settings}
\label{settings}
We use a $d = 25$ pose keypoints as extracted from OpenPose. Empirically, the value of $\alpha_c$ is set to $0.25$ for all our following experiments. The pose completion network is trained with the ADAM optimizer \cite{kingma2014adam} with learning rate $10^{-3}$ and dropout $0.5$ applied to the input layer. The local stream network is trained with ADAM too with learning rate of $10^{-4}$ and $N_\text{local} = 4$. The global stream network is trained on the residuals between consecutive frames with ADAM at learning rate $0.001$ and $N_\text{global} = 2$. We use Keypoint Displacement Error (KDE) defined as the average $\ell_2$ distance between the predicted and the ground truth keypoints to measure model's 
performance, defined as $
\text{KDE} = \nicefrac{(\sum_{j=t+1}^{t+ t_f}||\hat{p}_j - {p}_j||_1)}{t_f}.
$
\noindent Naturally, Mean KDE measures the Keypoint displacement error per human joint. For global stream, we use the same formulation but with $d=1$ in contrast to $d=25$ for local stream.  
\begin{table}[]
\caption{Average KDE for several popular sequence-to-sequence models, baselines and our proposed method. Despite their success at modelling other sequence learning tasks, a direct approach in predicting human locomotion in egocentric view with these models fail due to the complicated interaction between the local and global structures.}\label{tab:withooutdecomp}
\centering  \vspace{-6pt}
\begin{tabular}{lcc}
\hline
    \textbf{Method} & \textbf{Decomposition} & \textbf{KDE} \\ \hline
    Last Observed-Velocity & - & 195.5 \\ 
    Constant-Velocity & - & 48.9 \\
    Zero-Velocity & - & 40.4 \\
    LSTM-ED \cite{sutskever2014sequence} & \xmark & 24.4 \\
    TCNN \cite{lea2017temporal} & \xmark & 31.8 \\ %
    GRU-ED \cite{cho2014learning} & \xmark & 23.8 \\
    Structural RNN \cite{jain2016structural} & \xmark & 31.7 \\
    \hline 
    Ours & \cmark & \textbf{10.9} \\
    \hline
\end{tabular}
\label{tab: results}
\end{table}
\begin{table}[t]
\caption{Ablation Study on our method showing the relative importance of different components.}\label{tab: ablation}
\centering  \vspace{-6pt}
\begin{tabular}{lccc}
\hline
    \textbf{Stream} & \textbf{Completion}  & \textbf{Disentanglement} & \textbf{KDE} \\ \hline
    Global & \xmark & - & 6.8 \\
    Global & \cmark & - & 5.6 \\
    Local& \xmark & -  &9.5 \\
    Local & \cmark & - & 6.3 \\
    \hline
    Both & \xmark & \xmark & 22.1 \\
    Both & \cmark &\xmark & 18.2 \\
    Both & \xmark & \cmark & 15.4 \\
    Both & \cmark & \cmark &\textbf{10.9} \\
    \hline
\end{tabular}
\label{tab: results}
\vspace{-5mm}
\end{table}
\subsection{Pose Completion Results}

\label{posecompletion_results}
As described in Section \ref{posecompletion}, we use a symmetrical over complete auto-encoder for pre-processing noisy detections. Our model comprises of a compressing module that encodes the $2d = 50$ dimensional pose into $d_{ae} = 10$ dimensions, in three equally proportionate non linear layers. Symmetrically, the decoder decompresses the latent pose to the original $50$ dimensional space. A dropout of $0.5$ dropout is applied to input layer and training is done only on poses with all keypoints having confidence more than $c_t$. This is trained as mentioned in Section \ref{settings} to yield $9.6$ unweighed Mean KDE on high confidence pose detections on JAAD dataset. In practice, this means that on average the module is able to predict a missing joint within an $\ell_1$ ball of radius $10$ pixels centered at the joint's true location in a frame. Considering that the average resolution of frames in JAAD is $1408 \times 790$ pixels and the pose completion network is not trained to regress on residuals but instead predicts the absolute co-ordinates in original frame resolution, this means that the error is about $1\%-1.5\%$ of the frame dimensions and quite tolerable for our forecasting task. 
\begin{figure*}
\begin{center}
\includegraphics[width = \textwidth]{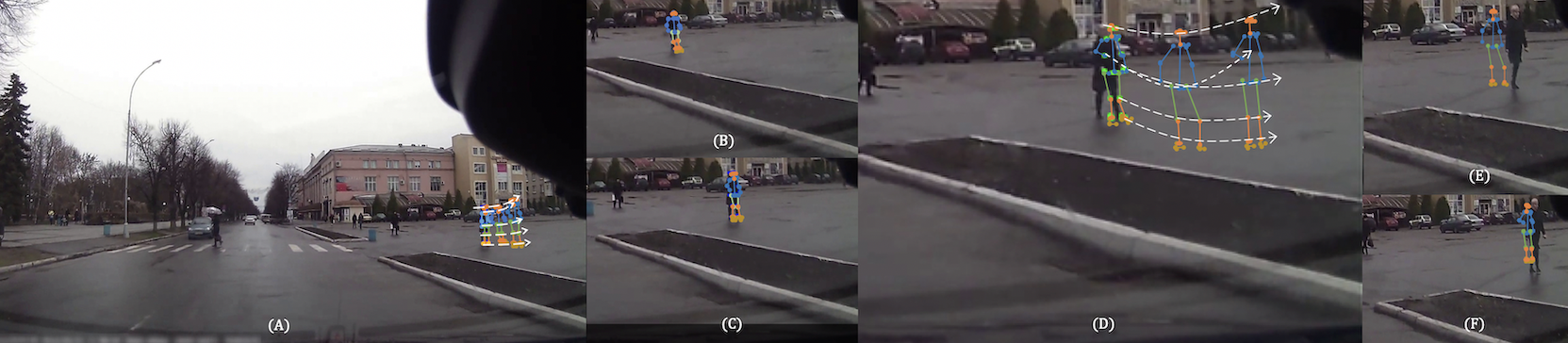}
\end{center}  \vspace{-15pt}
\caption{\small{Qualitative results from the pose prediction module. (A) shows the $t_p = 15$ length input sequence of poses for a pedestrian walking on the sidewalk. (B) and (C) show the cropped frames with the pedestrian and the corresponding filled in pose at the start and end of the input sequence. (D) shows the prediction pedestrian locomotion for the next $t_f = 15$ frames. (D), (E), and (F) also show the predicted poses at the start, intermediate, and end of the output sequence respectively. Note that the actual position of the pedestrian represents the ground-truth in (D), (E) and (F). More visualizations of our predictions are in the appendix.}}
\label{fig: qual_results}
\vspace{-6mm}
\end{figure*} 
\begin{table}[]
\caption{KDE comparison of our method with GRU-ED (best performing baseline from Table \ref{tab:withooutdecomp}) for different output time horizons for the global stream with input fixed at $15$ frames. For interpretability, $t_f = 15$ corresponds to half a second into the future.}
\label{tab: global_results}
\centering  \vspace{-6pt}
\begin{tabular}{@{}ccccccc@{}}
\toprule
$t_f$            & 5     & 10     & 15 & 20   & 25 & 30     \\ \midrule
GRU-ED &  5.7 & 6.6 & 9.7 & 11.4  & 14.1 & 18.3  \\
Our Method & 4.3 & 5.1  & 5.6  & 7.8 & 10.4  & 11.4    \\
\bottomrule
\end{tabular}
\vspace{-3mm}
\end{table}

\subsection{Stream Forecasting}
\label{final_results}
We compare our proposed method with several other baselines trained from scratch with the same input (\ie uncompleted pose estimates from Openpose) and output data. These methods are briefly described below:

\noindent{\textbf{Encoder Decoder Architectures}} : We train the popular Encoder Decoder architecture for sequence-to-sequence learning with several backbones like LSTM \cite{hochreiter1997long, sutskever2014sequence}, Gated Recurrent Units \cite{cho2014learning} and recently introduced Quasi RNN \cite{bradbury2016quasi}. All the models are trained from scratch on the same input data and confidence values but without any explicit spatial encoding. Also, Quasi RNN is the most parallelizable and train much faster than other recurrent counterparts. Hence, it forms the backbone of our method. 

\noindent{\textbf{Temporal Convolutional Neural Networks}}:  Lea et al. \cite{lea2017temporal} recently proposed Temporal Convolutional Networks which have also shown promising performance on a variety of sequence-to-sequence tasks. We found that TCNN trained with the same input signals, are the fastest to train and perform the best among deep learning methods without decomposition but are still much worse than the simple zero-velocity baseline. While TCNN perform well before decomposition, the future time horizon length $t_f$ is constrained by the input sequence length $t_p$ and does not allow flexible predictions as with other recurrent architectures. 

\noindent{\textbf{Strutural RNN}}\cite{jain2016structural}: SRNN has also achieved state of the art performance on several sequence prediction tasks including pose forecasting. However, we observe that its capacity is much larger than other sequence-to-sequence models because of larger number of trainable parameters that does not generalize well to our task (limited data regime).

 \noindent{\textbf{Uniform Velocity}}: We also compare our method with simple yet powerful baselines \cite{ma2017forecasting, scholler2019simpler} based on zero, constant, and last observed velocity assumptions. In Zero Velocity, the future is forecasted assuming no movement from the last observed position. Similarly, in constant-velocity/last-observed-velocity, the forecasted velocity is assumed to be identically equal to the average observed velocity/last-observed-velocity and this assumption is rolled out in time to predict future positions. 
 
 \noindent{\textbf{Results Discussion}}: Table \ref{tab:withooutdecomp} reports the Average KDE achieved by these approaches for predicting the pedestrian locomotion for $t_f = 15$ frames into the future given the previous $t_p = 15$ frames. Our method performs better than several widely used sequence-to-sequence models and also state-of-the-art pose forecasting models like Structural RNN \cite{jain2016structural}. Figure \ref{fig: qual_results} shows some qualitative results from our method for pedestrian locomotion prediction.

\noindent\textbf{Ablation}: We perform the ablation study on our proposed method to investigate the importance of each component. As Table \ref{tab: ablation} indicates, both disentanglement and completion are quite central to our method. Also, the significant increase in performance conferred by decomposing the streams confirms our hypothesis that disentangling is essential to locomotion forecasting and reduces the complication of overlapping simultaneous movements. Also noteworthy is the decrease in performance due decomposing without pose completion. This backs up our proposition that completion is essential to disentangling and data is subject to severe method-induced noise without pose completion.

\noindent\textbf{Time Horizon Experiments}: We extensively test our novel egocentric global stream 
(trajectory) forecasting module for various time horizons against the best performing baseline from Table \ref{tab: results}. Table \ref{tab: global_results} shows these trends for time horizons from $165$ ms to $1$ second. We observe that as the horizon increases and the trajectories become longer, our method outperforms the baseline by an increasing and larger margin. This is because our model accounts for the camera motion, which can accumulate over longer time horizons leading to a large error in egocentric motion agnostic methods. 
\vspace{-4mm}
\section{Discussion}
\noindent{Taking} into consideration the results discussed in the last section, we would like to revisit our original hypothesis and discuss final remarks that also highlight the new findings.

\noindent\textbf{Disentanglement}: We hypothesize that disentangling the human dynamics into overlapping but simpler streams helps overall prediction. This is corroborated with the results and ablation experiments discussed before. We further suspect that disentanglement performed in our proposed manner is useful in predicting motion dynamics of any object, cognizant or inanimate. The overall prediction difficulty of the spatially disentangled parts is lesser than that of the whole.

\noindent\textbf{Finetuning Pose Detection}: The state of the art pose detection network is pretrained on general purpose pose datasets and is not adapted to this setting. The pose completion network finetunes these detection conditioning on the information that the majority of human poses for pedestrian locomotion prediction are upright, as is the case when pedestrian is standing, walking or running. Thus, it implicitly also works as an adaptation step to use pose regularities specific to the setting of locomotion prediction. Additionally, it fills in the missing data because as discussed in Section \ref{posecompletion}, it is the existence of the keypoints and not so much their accuracy that helps the decomposition. 

\section{Conclusion}
We propose a human locomotion forecasting pipeline that disentangles the complex pedestrian locomotion into simpler tasks of global and local motion prediction. We also present a pose completion module and an Encoder-Recurrent-Decoder style pose prediction network. The pose completion module disentangles the overlapping motion and fills in the missing detected joints. Furthermore, the pose prediction network forecasts these streams separately which are finally merged for the final prediction. We demonstrate that our method improves the overall forecasting performance and achieves state-of-the art short term human locomotion forecasting in egocentric view.   
\section*{Acknowledgments}
Toyota Research Institute (TRI) provided funds to assist the authors with their research but this article solely reflects the opinions and conclusions of its authors and not TRI or any other Toyota entity.

{\small
\bibliographystyle{ieee}
\bibliography{main}
}
\newpage
\begin{figure*}
\begin{center} 
\includegraphics[width = \textwidth]{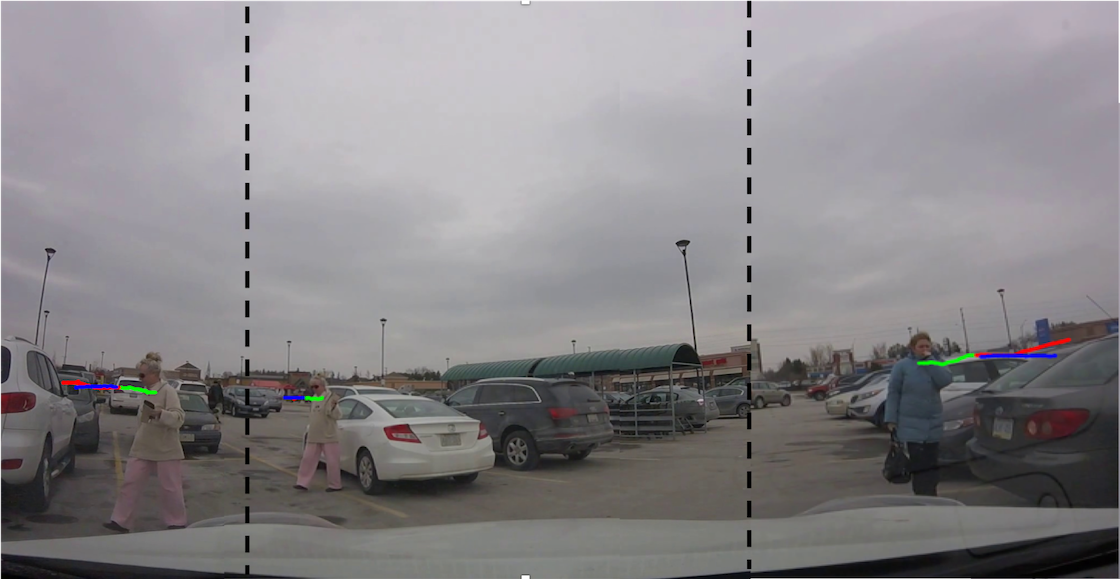}
\caption{Visualization as a collage for three different trajectory prediction instances (in the same video). Green denotes the input to the network, Blue is the prediction and red is the ground truth. The trajectories are also a bit distorted from typical walking behaviour because of the car's (camera's) motion itself.}
\end{center}
\end{figure*}
\begin{figure*}
\begin{center} 
\includegraphics[width = \textwidth]{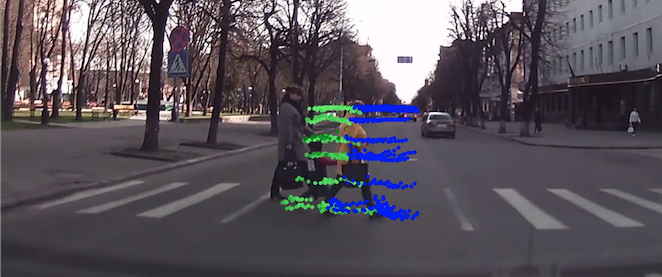}
\caption{Visualization of locomotion prediction (trajectory + pose) for a \underline{typical} walking pedestrian (cropped). The green dots indicate the historical input to the whole module and the blue denote the predicted output. Note how the large swinging step motion of legs and the smaller jiggle of the arms are accurately predicted.}
\end{center}
\end{figure*}
\begin{figure*}
\begin{center} 
\includegraphics[width = \textwidth]{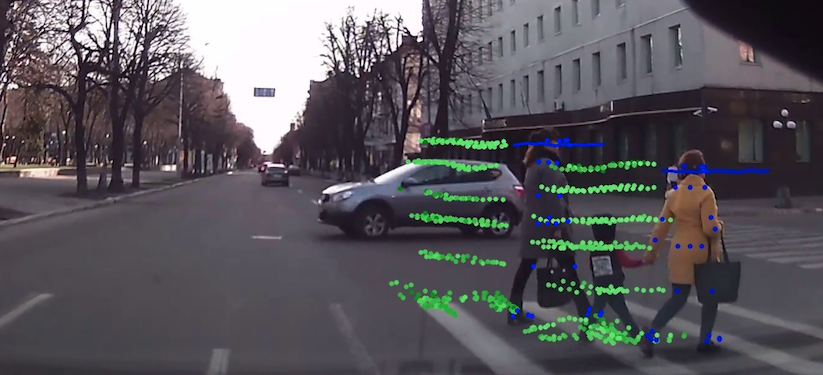}
\caption{Visualizing complete predicted trajectory and instantaneous predicted pose for two pedestrians. Green dots represent the previous history (distorted due to camera motion since their observance), blue represents predictions. The predicted pose closely corresponds to the future ground truth, as can be verified visually. Full length video visualized with predictions can be accessed at: \href{https://youtu.be/-4gCHoSrWvY}{https://youtu.be/-4gCHoSrWvY}}
\end{center}
\end{figure*}
\begin{figure*}
\begin{center} 
\includegraphics[width = \textwidth]{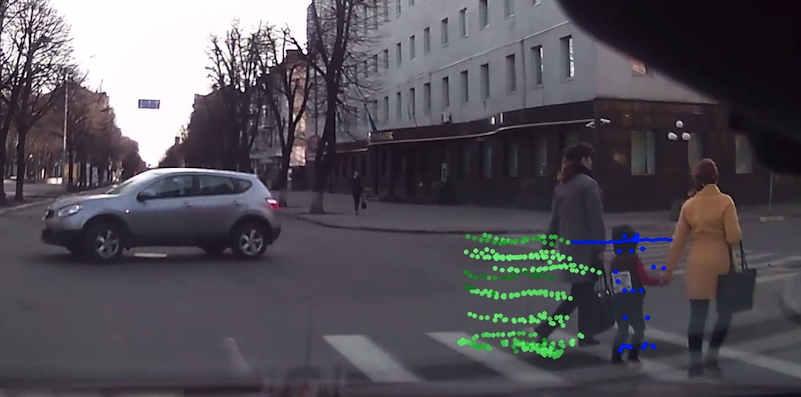}
\caption{Another visualization of trajectories with instantaneous poses showing the predictions on the child as well. Following the same convention for green as input and blue being the predictions, notice that the results for child are reasonably good as well since we do not impose arbitrary human priors on human height for depth estimation etc.}
\end{center}
\end{figure*}
\end{document}